\pdfoutput=1

\documentclass[11pt]{article}

\usepackage[final]{acl}

\usepackage{times}
\usepackage{latexsym}
\usepackage{booktabs}
\usepackage{array}
\usepackage{booktabs}  
\usepackage{makecell}
\usepackage{colortbl}
\usepackage{graphicx}
\usepackage{subcaption}
\usepackage{dcolumn} 
\usepackage{multirow}
\usepackage{color}
\usepackage{tabularx}
\usepackage{stfloats} 
\usepackage{listings}
\usepackage{threeparttable}
\usepackage{amsmath}  
\usepackage{amsthm}   
\usepackage{amssymb}  
\usepackage{xurl}
\usepackage{longtable}
\usepackage{adjustbox}
\usepackage{threeparttable}
\usepackage{booktabs}
\usepackage{hhline}
\usepackage{xcolor} 
\usepackage{color,colortbl} 
\usepackage{booktabs}
\usepackage{graphicx}

\definecolor{Gray}{gray}{0.92}
\usepackage[T1]{fontenc}

\usepackage[utf8]{inputenc}

\usepackage{microtype}

\usepackage{inconsolata}

\usepackage{graphicx}
\usepackage{tcolorbox}

\PassOptionsToPackage{prologue,dvipsnames}{xcolor}
\newcommand{\ours}{ThinkPatterns-21k}

\definecolor{monologuecolor}{HTML}{D4DCF0} 
\definecolor{decompositioncolor}{HTML}{E3F0D8}
\definecolor{askcolor}{HTML}{FDF3CF} 
\definecolor{debatecolor}{HTML}{FEF0E7}
\definecolor{criticcolor}{HTML}{F5D7D9} 

\newcommand{\monologue}[1]{
    \mbox{\raisebox{-3pt}{\tcbox[colback=monologuecolor,colframe=black,size=small,boxsep=1pt]{\texttt{#1}}}}
}

\newcommand{\decomposition}[1]{
    \mbox{\raisebox{-3pt}{\tcbox[colback=decompositioncolor,colframe=black,size=small,boxsep=1pt]{\texttt{#1}}}}
}

\newcommand{\ask}[1]{
    \mbox{\raisebox{-3pt}{\tcbox[colback=askcolor,colframe=black,size=small,boxsep=1pt]{\texttt{#1}}}}
}

\newcommand{\debate}[1]{
    \mbox{\raisebox{-3pt}{\tcbox[colback=debatecolor,colframe=black,size=small,boxsep=1pt]{\texttt{#1}}}}
}

\newcommand{\critic}[1]{
    \mbox{\raisebox{-3pt}{\tcbox[colback=criticcolor,colframe=black,size=small,boxsep=1pt]{\texttt{#1}}}}
}

\title{\ours:
A Systematic Study on the Impact of Thinking Patterns in LLMs}

\author{
Pengcheng Wen$^1$, 
Jiaming Ji$^2$, 
Chi-Min Chan$^1$, 
Juntao Dai$^3$ \\  
\textbf{Donghai Hong$^2$, Yaodong Yang$^2$,
Sirui Han$^1$\thanks{~~Corresponding author.}, 
Yike Guo$^{1*}$ }
\\
 $^1$Hong Kong University of Science and Technology \\
 $^2$Peking University \\
  $^3$Zhejiang University \\
  \texttt{pc.wen@connect.ust.hk}  \\
}

\begin{document}
\maketitle

\begin{abstract}
Large language models (LLMs) have demonstrated enhanced performance through the \textit{Thinking then Responding} paradigm, where models generate internal thoughts before final responses (aka, System 2 thinking). However, existing research lacks a systematic understanding of the mechanisms underlying how thinking patterns affect performance across model sizes.
In this work, we conduct a comprehensive analysis of the impact of various thinking types on model performance and introduce~\ours, a curated dataset comprising 21k instruction-response pairs (QA) collected from existing instruction-following datasets with five thinking types. 
For each pair, we augment it with five distinct internal thinking patterns: one unstructured thinking (monologue) and four structured variants (decomposition, self-ask, self-debate and self-critic), while maintaining the same instruction and response. Through extensive evaluation across different model sizes (3B-32B parameters), we have two key findings: (1) smaller models (<30B parameters) can benefit from most of structured thinking patterns, while larger models (32B) with structured thinking like decomposition would degrade performance and (2) unstructured   monologue demonstrates broad effectiveness across different model sizes. Finally, we released all of our datasets, checkpoints, training logs of diverse thinking patterns to reproducibility, aiming to facilitate further research in this direction.

\end{abstract}

\section{Introduction}
\begin{figure}[t]
\includegraphics[width=\columnwidth]{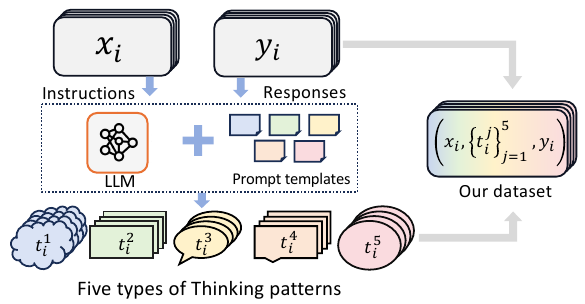}
\caption{\textbf{Overview of our dataset construction pipeline.} The process begins with gathering instruction-response pairs. These are subsequently processed by a large language model with specific prompts to produce five internal thinking patterns: \monologue{monologue}, \decomposition{decomposition}, \ask{self-ask}, \debate{self-debate}, and \critic{self-critic}, which are then merged to create our final dataset.}
\label{fig:pipeline}
\end{figure}

Large language models (LLMs) have traditionally been designed to generate responses immediately upon receiving human's instruction \citep{openai2023gpt4, team2024gemini, touvron2023llama}. However, the emergence of reasoning models like OpenAI's O1/O3~\cite{jaech2024openai}, QwQ~\cite{qwq-32b-preview} and Deepseek R1~\cite{guo2025deepseek} have demonstrated the effectiveness of a new paradigm: \textit{Thinking then Responding}, also know as system 2 thinking~\cite{evans2003two}. 
Under this paradigm, models engage in explicit thinking processes before final response. This approach has shown remarkable success, achieving or even surpassing human-level performance in various domains, including mathematical problem-solving \citep{guan2025rstar, zhang2025lessons}, coding tasks \citep{zhang2024o1}, and etc~\citep{zhong2024evaluationopenaio1opportunities}.

However, a fundamental challenge exists in the current landscape: traditional LLMs are not inherently designed to follow the \textit{Thinking then Responding} paradigm and require specialized training to induce such behavior. 
Specifically, existing training data often omits the thinking process entirely~\cite{zelikman2024quietstar} and lacks systematic annotation of thinking patterns that could guide model reasoning~\cite{wu2024thinkingllmsgeneralinstruction}. Recent works have explored several technical approaches to address the challenge. 
One line of these works focuses on synthetic data generation, including using Monte Carlo Tree Search for reasoning path synthesis \cite{qin2024o1replicationjourneystrategic,guan2025rstar}, leveraging GPT-4o to generate thinking demonstrations \cite{chen2024huatuogpto1medicalcomplexreasoning}, and distilling from existing reasoning models \cite{guo2025deepseek,muennighoff2025s1}. This approach directly addresses the limitation of lacking thinking process annotations in training data. 

The other line employs reinforcement learning (RL)~ \cite{sutton2018reinforcement}, exemplified by Deepseek R1 Zero \cite{guan2025rstar,kimiteam2025kimik15scalingreinforcement}, which enables models to learn reasoning paths autonomously without 
 explicit thinking process annotations. However, these approaches still face significant limitations. 
 First, RL-based methods typically generate unstructured thinking patterns, and existing synthetic data generation approaches have also focused on unstructured thinking patterns. 
 Given the demonstrated effectiveness of structured prompting techniques (e.g., self-ask~\cite{press-etal-2023-measuring}, debate~\cite{liang-etal-2024-encouraging}, divide and conquer~\cite{wang-etal-2023-plan,khot2023decomposed}) in enhancing model performance, we believe it is promising to enable models to follow structured thinking patterns during reasoning. 
 Furthermore, there remains a limited understanding of how different thinking patterns perform across model sizes, making it challenging to determine optimal thinking patterns without extensive experimentation.

To overcome these limitations, we conduct a comprehensive study on the impact of different thinking patterns on LLM performance across various model sizes. 
As Figure \ref{fig:pipeline}, we curate a large-scale dataset containing 21k instruction-response pairs,  augmenting each pair with five distinct thinking patterns -- one unstructured thinking (monologue) and four structured variants (decomposition, self-ask, self-debate and self-critic). Notably, we preserve the same instruction and response while varying internal thinking processes for each pattern. 
Through systematic experiments on Arena Hard \cite{li2024crowdsourced} and Alpaca Eval 2 \cite{alpaca_eval} benchmarks across various model sizes (3B-32B parameters), \textbf{ we find that smaller models (<30B parameters) can benefit from most structured thinking patterns, while larger models (32B parameters) perform better with unstructured thinking. Remarkably, unstructured   monologue shows consistent effectiveness across most model sizes.} 
Our contributions can be summarized as:
\begin{itemize}
\item We introduce \ours, a large-scale dataset containing 21k instruction-response pairs, each augmented with five distinct internal thinking patterns (one unstructured and four structured variants), aiming to facilitate further research in this area.   
\item We present the first comprehensive study on the interaction between model sizes and thinking patterns, demonstrating that smaller models (<30B parameters) can benefit from most structured thinking patterns, while larger models (32B parameters) perform better with unstructured thinking. Our analysis also reveals that unstructured monologue shows consistent effectiveness across most model sizes.
\end{itemize}

\begin{figure*}[ht]
  \includegraphics[width=\textwidth]{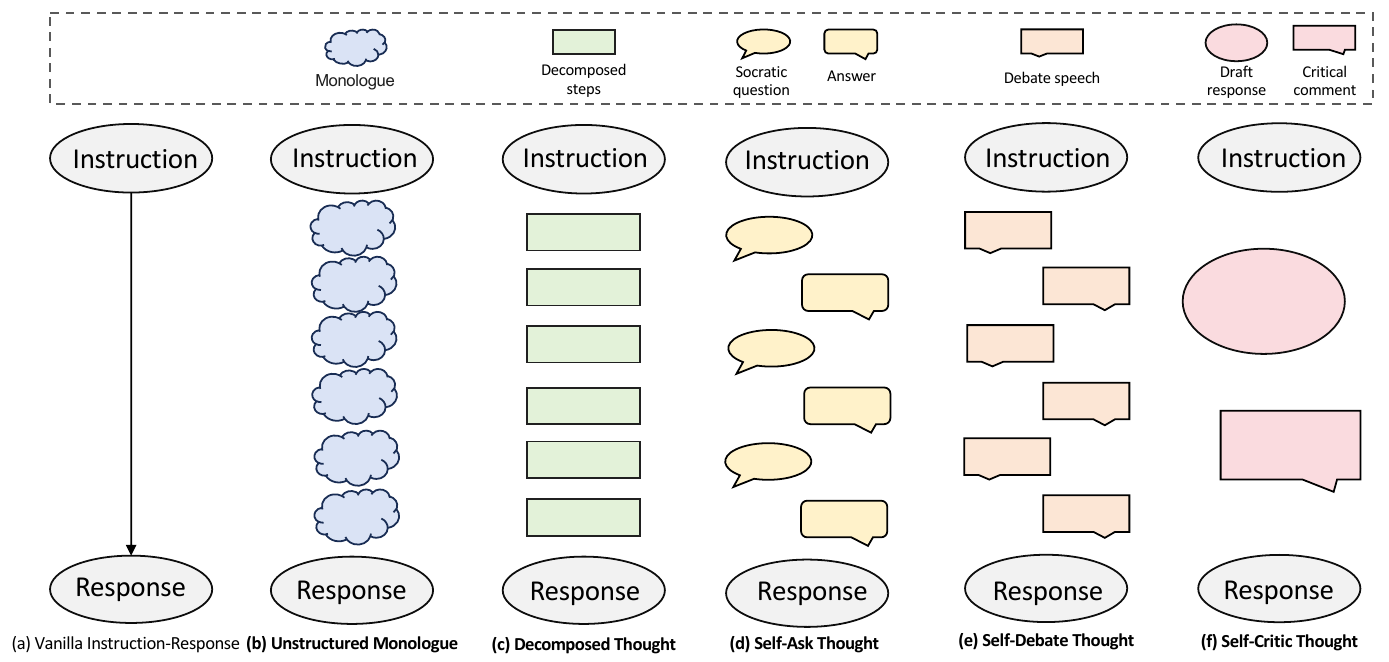}
  \caption{\textbf{Comparison of Different Internal Thinking Patterns.} This figure illustrates the contrast between vanilla instruction-response paradigm and our proposed five internal thought types: (a) vanilla instruction-response  \hspace{2em}  \textbf{(b) Unstructured \monologue{Monologue}}, mimicking natural human internal monologue \textbf{(c) \decomposition{Decomposition} Thought}, which systematically breaks down complex tasks into manageable sub-problems \textbf{(d) \ask{Self-Ask} Thought}, implementing Socratic questioning for deeper exploration \textbf{(e) \debate{Self-Debate} Thought}, facilitating internal debate dialogue to reach optimal solutions \textbf{(f) \critic{Self-Critic} Thought}, incorporating self-evaluation and refinement mechanisms. Each thought pattern demonstrates a unique reasoning pathway and problem-solving strategy.}
  \label{fig:comparison}
\end{figure*}

\section{Related Work}

\paragraph{Prompt Engineering for Reasoning}
The development of language model reasoning has witnessed significant progress through prompt engineering innovations. Following the seminal work of Chain-of-Thought(CoT)~\cite{wei2022chain}, researchers have proposed diverse thinking frameworks. These include Self-Ask \cite{press-etal-2023-measuring} for interactive problem decomposition, Self-Debate \cite{liang-etal-2024-encouraging} for dialectical reasoning, and Divide-and-Conquer \cite{wang-etal-2023-plan,khot2023decomposed} for systematic problem solving. However, these approaches merely focus on prompt engineering without internalizing these thinking patterns into the model's parameters.
\paragraph{Model Training for Enhanced Reasoning}
Complementing prompt engineering efforts, another research direction focuses on enhancing reasoning capabilities through data synthesis and model training. Recent advances include the application of Monte Carlo Tree Search for reasoning path generation \cite{qin2024o1replicationjourneystrategic,guan2025rstar} and the utilization of GPT-4 for demonstration synthesis \cite{chen2024huatuogpto1medicalcomplexreasoning}. Knowledge distillation from advanced reasoning models has also shown promise \cite{min2024imitateexploreselfimprovereproduction,guo2025deepseek,sky_t1_2025,Open_Thoughts,muennighoff2025s1,ye2025limoreasoning,xu2025redstardoesscalinglongcot}. While these approaches have shown promising results, they have not thoroughly investigated the advantages of structured thinking patterns, and the relationship between model size and thinking pattern effectiveness remains largely unexplored.
\paragraph{Test Time Scaling for LLMs}
The application of test-time compute scaling has emerged as a promising approach to enhance the reasoning capabilities of large language models. Broadly, this methodology can be divided into two strategies: generating a single, extended CoT and performing repeated sampling of CoT responses. In the single long CoT approach, models—such as OpenAI's O1/O3, QwQ and Deepseek R1 trained to produce individual, lengthy CoT sequences that incorporate in-context reflection and backtracking to address complex reasoning tasks ~\citep{guo2025deepseek,jaech2024openai,qwq-32b-preview}. Alternatively, repeated sampling techniques (e.g., the Best-of-N strategy or search-guided mechanisms like MCTS) enhance reasoning performance by generating multiple candidate responses, often guided by search algorithms or reward models, and then selecting the most effective outcome~\citep{snell2024scaling,brown2024large}. In this work, we focus specifically on the single, extended CoT approach and explore different structural variants of single CoT.

\section{Dataset}
In this section, we introduce our \ours~dataset and describe its key specifications.

\vspace{0.3cm}

\paragraph{\ours.}
Let $\mathcal{C} = \{(x_i, y_i)\}_{i=1}^N$ denote our seed dataset, comprising $N=21$K instruction-response pairs, with $x_i$ and $y_i$ representing the instruction and its corresponding response, respectively.
For each pair $(x_i, y_i)$, we augment it with a thinking pattern set $\mathcal{T}$.

\begin{align*}
\mathcal{T} &= \{t^1, t^2, t^3, t^4, t^5\},
\end{align*}

\vspace{0.3cm}

where $t_1$, unstructured monologue; $t_2$, decomposition thought; $t_3$, self-ask thought; $t_4$, self-debate thought; $t_5$, self-critic thought.
\vspace{0.3cm}

Thus, Our final dataset~\ours~can be represented as,
\vspace{0.3cm}
\begin{align*}
    \mathcal{D} = \{(x_i, \{t_i^j\}_{j=1}^{|\mathcal{T}|}, y_i)\}_{i=1}^N.
\end{align*}

\subsection{Thinking patterns in set \texorpdfstring{$\mathcal{T}$}{T}}
Our dataset contains five thinking patterns. Below we introduce each thinking pattern in detail. See Appendix \ref{sec:appendix} for concrete examples.

\begin{figure}[h]
  \includegraphics[width=\linewidth]{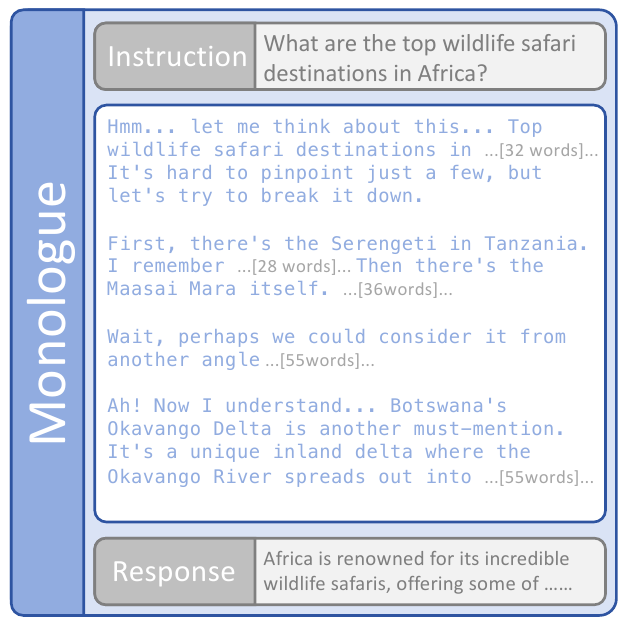}
  \caption{\textbf{Example of  \textbf{Unstructured \monologue{Monologue}}.} }
  \label{fig:monlogue}
\end{figure}
\textbf{Unstructured Monologue}
represents a free-form thinking pattern without explicit structural constraints, as shown in Figure \ref{fig:monlogue}. Most current reasoning models, such as OpenAI's O1 series~\cite{jaech2024openai} and Deepseek R1~\cite{guo2025deepseek}, follow this unstructured monologue thinking pattern to generate their reasoning processes.
\begin{figure}[h]
  \includegraphics[width=\linewidth]{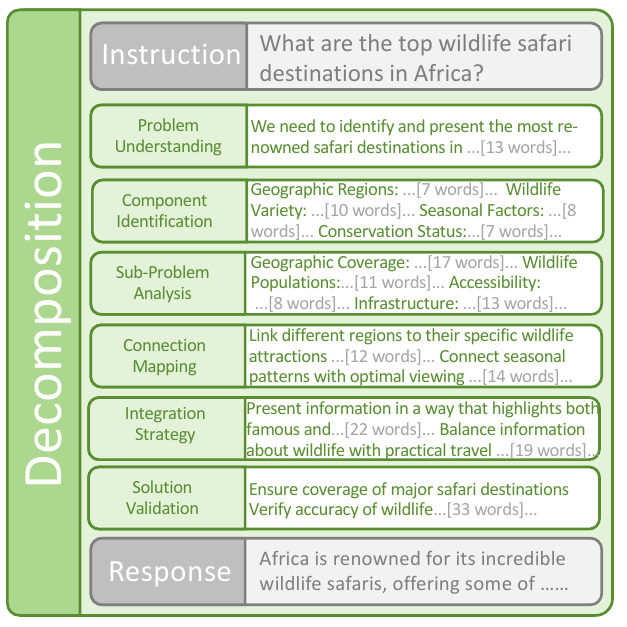}
  \caption{\textbf{Example of \textbf{\decomposition{Decomposition} Thought}.} }
  \label{fig:decomposed}
\end{figure}

\vspace{1em}
\textbf{Decomposition Thought} is a systematic and structured thinking framework that adopts divide-and-conquer principles~\cite{wang-etal-2023-plan,khot2023decomposed}. It systematically breaks down complex problems into smaller, manageable components and solves them iteratively, enabling more transparent and trackable reasoning processes in five steps, as illustrated in Figure \ref{fig:decomposed}. This decomposition strategy not only enhances problem-solving efficiency but also improves the interpretability of the reasoning path.
 
\begin{figure}[h]
  \includegraphics[width=\linewidth]{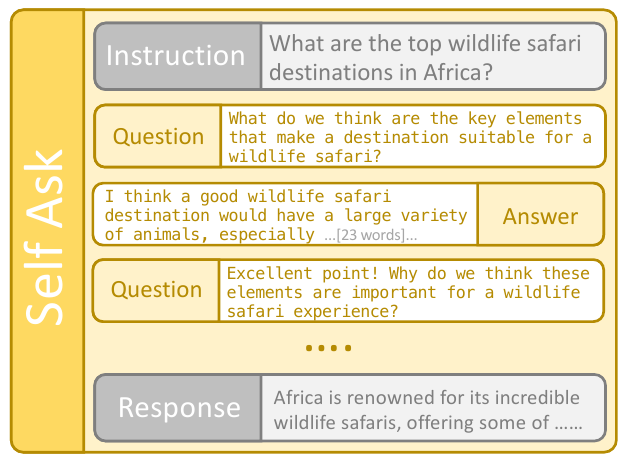}
  \caption{\textbf{Example of \textbf{\ask{Self-Ask} Thought}.}}
  \label{fig:ask}
\end{figure}

\textbf{Self-Ask Thought} is another potentially effective thought, inspired by Socratic questioning workflow~\cite{press-etal-2023-measuring}. As shown in Figure \ref{fig:ask}, under this type of thought, the model iteratively raises sub-questions from multiple angles based on the user's instruction and answers them itself. By engaging in this recursive questioning pattern, the model can explore problems more deeply, identify potential blind spots, and generate more thorough and well-reasoned responses.

\begin{figure}[h]
  \includegraphics[width=\linewidth]{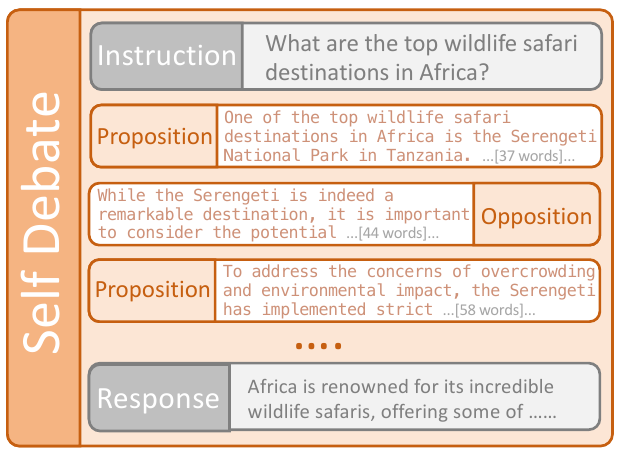}
  \caption{\textbf{Example of \textbf{\debate{Self-Debate} Thought.}} }
  \label{fig:debate}
\end{figure}

\textbf{Self-Debate Thought}
is a reasoning framework that enables models to conduct structured internal debates by adopting opposing perspectives within a single-model reasoning process. Drawing inspiration from multi-agent debate approaches \cite{chan2024chateval,liang-etal-2024-encouraging}, which have shown enhanced performance through structured argumentation, our mechanism (illustrated in Figure \ref{fig:debate}) internalizes these benefits through systematic self-dialogue. This approach allows models to leverage the advantages of debate-style reasoning without requiring multiple distinct agents, effectively transforming external multi-agent debates into an internal reasoning process.

\begin{figure}[h]
  \includegraphics[width=\linewidth]{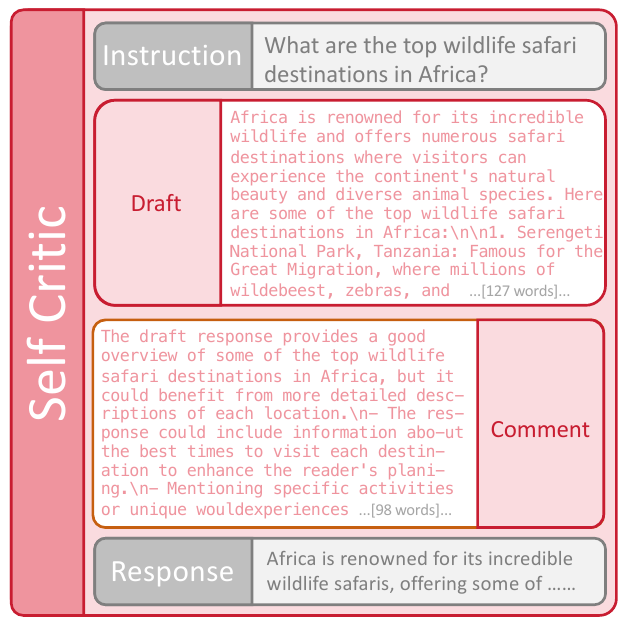}
  \caption{\textbf{Example of \textbf{\critic{Self-Critic} Thought.}}}
  \label{fig:cirtic}
\end{figure}

\textbf{Self-Critic Thought} is an iterative reasoning framework where models engage in self-comment and refinement of their initial responses, as illustrated in Figure \ref{fig:cirtic}. In this process, the model first generates a draft answer, then critically evaluates its own response by providing detailed comments, before producing a refined final output. Previous research has demonstrated that such self-commenting and revision workflows can significantly enhance model performance \cite{madaan2023selfrefineiterativerefinementselffeedback}. Through this systematic self-review mechanism, we aim to improve the quality and reliability of model responses by identifying and addressing potential weaknesses in the initial draft.

\subsection{Dataset Construction Process}

\paragraph{Instruction-Response pairs collection}
To construct our dataset with explicit internal thought processes, we first sampled instruction-response pairs from existing instruction following datasets. We selected the Infinite Instruct dataset\footnote{\url{https://huggingface.co/datasets/BAAI/Infinity-Instruct}} as our seed dataset due to its comprehensive coverage of diverse instruction-following scenarios. After filtering to retain only English data, we randomly sampled 21k instruction-response pairs while maintaining a balanced distribution across different task categories (as shown in Figure \ref{fig:categories}).

\paragraph{Thinking Patterns Construction.}
To generate internal thinking patterns for each instruction-response pair, we employed GPT-4o as our primary annotation model. We designed specific prompts (provided in Appendix \ref{sec:appendix}) to guide GPT-4o~\citep{hurst2024gpt} in generating internal thoughts that align with our five designed thinking patterns. The prompts were carefully crafted to ensure the generated thoughts reflect natural cognitive progression, maintain coherence between the thinking process and final response, and cover diverse aspects of problem-solving relevant to each task type.

For quality validation, three annotators performed spot checks on a sample of the generated internal thoughts. The review examined the logical coherence of the reasoning chain and its relevance to the original instructions. This sampling-based verification helped ensure the basic quality of our thought simulation approach while maintaining reasonable instruction-response relationships.
\vspace{2em}

\begin{table}[t!]
\centering
\caption{\textbf{Statistical Metrics of Thinking Patterns.} This table presents a comparative analysis of different thinking patterns in terms of their average token length ($\overline{|Tokens|}$) and average number of turns ($\overline{Turns}$). Note that \monologue{monologue} is inherently a single-step process (1 turn), \decomposition{decomposition} is designed with a fixed 5-step structure, and \critic{self-critic} follows a predetermined 2-step process. In contrast, \ask{self-ask} and \debate{self-debate} exhibit variable turn counts based on their iterative nature.}
\label{tab:response-gen}
\resizebox{\columnwidth}{!}{
\begin{tabular}{l|ccc}
\toprule
\textbf{Thinking Patterns} & \textbf{$\overline{|Tokens|}$} & \textbf{$\overline{Turns}$} & \textbf{Pattern} \\
\midrule
 \monologue{Monologue}  & 1300.50 & \boxed{1} & Single-pass \\
\decomposition{Decomposition}  & 1218.28 & \boxed{5} & Multi-step \\
\ask{Self-Ask} & 1393.09 & 36.51 & Iterative \\
\debate{Self-Debate}  & 1486.97 & 24.38 & Dialectical \\
\critic{Self-Critic}  & 1445.12 & \boxed{2} & Review-based \\
\bottomrule
\end{tabular}
}
\label{tab:statis}
\end{table}
\paragraph{Statistical Analysis of Thinking Patterns}
Table \ref{tab:statis} presents a comprehensive analysis of the five thinking patterns, revealing distinct characteristics in terms of their token length and interaction dynamics. The patterns exhibit notable variations in both their structural design and verbosity. self-debate demonstrates the highest token count ($\overline{|Tokens|}=1486.97$) among all patterns, reflecting its dialectical nature where multiple perspectives are thoroughly explored through an average of 24.38 turns. self-ask shows the highest average number of turns ($\overline{Turns}=36.51$), indicating its highly iterative question-answer approach, while maintaining a moderate token length ($\overline{|Tokens|}=1393.09$).
In contrast, monologue, decomposition, and self-critic follow more structured formats with predetermined turn counts. Monologue employs a single-pass approach, generating an average of 1300.50 tokens in one turn. Decomposition, despite its fixed 5-step structure, shows the most concise token usage ($\overline{|Tokens|}=1218.28$). Self-Critic combines initial response with subsequent comment in its two-step process, resulting in substantial token generation ($\overline{|Tokens|}=1445.12$).

This diversity in thinking patterns suggests their potential complementary roles: self-debate excels in comprehensive exploration, self-ask in progressive reasoning, decomposition in structured analysis, monologue in direct response, and self-critic in reflective thinking.

\begin{table*}[ht]
\centering
\resizebox{\linewidth}{!}{
\begin{tabular}{l||ccc|ccc|ccc|ccc}
    \toprule
    \bf{Models} $\rightarrow$ & \multicolumn{3}{c}{\bf{Qwen2.5-3B}} & \multicolumn{3}{c}{\bf{Qwen2.5-7B}} & \multicolumn{3}{c}{\bf{Qwen2.5-14B}} & \multicolumn{3}{c}{\bf{Qwen2.5-32B}}\\ 
    \midrule
     \bf{Benchmarks} $\rightarrow$& \multicolumn{2}{c}{\bf{AlpacaEval2}} & \bf{Arena-H}& \multicolumn{2}{c}{\bf{AlpacaEval2}} & \bf{Arena-H}& \multicolumn{2}{c}{\bf{AlpacaEval2}} & \bf{Arena-H}& \multicolumn{2}{c}{\bf{AlpacaEval2}} & \bf{Arena-H}\\ 
    \bf{Types} $\downarrow$ \bf{Metrics}$\rightarrow$ & \bf{LC}(\%) & WR (\%) & WR (\%)  & \bf{LC} (\%) & WR (\%) & WR (\%) & \bf{LC} (\%) & WR (\%) & WR (\%) & \bf{LC} (\%) & WR (\%) & WR (\%)\\ 
    \hhline{=||===|===|===|===}
    Instruction-Response  				& $\underline{11.49}$ 			& $\underline{9.24}$ 			& $23.40$ & $\underline{20.76}$ 			& $\underline{15.00}$ 			& $44.30$& $\underline{29.53}$ 			& $\underline{24.73}$ 			& $53.60$& $40.80$ 			& $39.11$ 			& ${68.00}$							\\
     \monologue{Monologue} 	& $\mathbf{24.70}$ 				& $\mathbf{25.31}$			& $\mathbf{32.50}$ 			&$25.28$ 				& $25.76$ 		&$\underline{31.20}$ 			& $\mathbf{44.68}$ 			& $\mathbf{43.04}$ 	& $\mathbf{66.20}$ 		&$\mathbf{47.87}$ 			& $44.91$ 			& $\mathbf{71.60}$ 				 \\
    \decomposition{Decomposition}  		& $18.66$ 			& $20.32$ 			& $\underline{21.90}$ 				& $32.26$ 			& $32.71$ 
    & $\mathbf{46.70}$ 			& $41.62$ 			& $36.94$  	& $58.40$ 
    & $\underline{35.59}$ 			& $\underline{24.71}$ 			& $\underline{50.80}$  			\\
    \ask{Self-Ask} 	& $21.65$ 			& $21.46$ 			& $26.40$	& $33.85$ 			& $30.06$ 
    & $46.60$			& $40.88$ 			& $37.40$ 	& $62.80$ 
    & $47.00$			& $41.92$ 			& $68.10$ 		\\
    \debate{Self-Debate}	& $21.97$ 			& $20.24$ 			& $25.20$	& $34.59$ 			& $31.57$ 
    & $42.40$			& $43.98$ 			& $40.67$ 		& $\underline{50.70}$ 
    & $45.90$			& $40.45$ 			& $70.60$ 	\\
    \critic{Self-Critic} 	& $21.88$ 			& $22.50$ 			& $30.30$	& $\mathbf{36.42}$ 			& $\mathbf{33.96}$ 
    & $39.20$			& $43.91$ 			& $41.49$ 		& $62.00$ 
    & $47.35$			& $\mathbf{45.50}$ 			& $71.40$ 		\\
    \bottomrule
\end{tabular}
}
\caption{\textbf{Performance comparison of Qwen2.5 models with different thinking patterns.} Results are reported on AlpacaEval2 (\textbf{L}ength \textbf{C}ontrol win rate and \textbf{W}in \textbf{R}ate) and Arena-Hard (\textbf{W}in \textbf{R}ate) benchmarks. Higher scores indicate better performance.}
\label{tab:main}
\end{table*}

\section{Experiment about Thinking Patterns}

\subsection{Experiment Setup}

To rigorously evaluate the effectiveness of our dataset, we conducted comprehensive experiments across multiple model sizes. This section details our experimental setup, evaluation methodology, and benchmarking approach.

\paragraph{Base Models}
To systematically evaluate across different model sizes, we selected base models ranging from 3B to 32B parameters in Qwen-2.5 series~\citep{qwen2.5}.

\paragraph{Evaluation Benchmarks}
We employed two widely-recognized benchmarks for evaluating open-ended language model performance:
\begin{itemize}
    \item \textbf{AlpacaEval 2} \citep{alpaca_eval}: Comprises 805 diverse instructions derived from real user interactions, providing a broad spectrum of practical scenarios. This benchmark uses GPT-4-Turbo (1106) as its baseline for comparative evaluation.
    
    \item \textbf{Arena-Hard} \citep{li2024crowdsourced}: Contains 500 deliberately challenging user queries designed to test advanced reasoning capabilities. The benchmark employs GPT-4-0314 as its evaluation baseline.
\end{itemize}

Both benchmarks utilize LLM-as-judge for automated comparative assessment, where model-generated responses are compared against baseline responses to determine relative performance.

\paragraph{Evaluation Metrics}
Our evaluation framework incorporates multiple complementary metrics:

\begin{itemize}
    \item \textbf{Win Rate (WR)}: Standard comparison metric measuring the proportion of responses preferred over the baseline.
    \item \textbf{Length-Controlled Win Rate (LC)}: For AlpacaEval 2, this metric normalizes for response length to minimize the length bias.
\end{itemize}

\paragraph{Instruction-Tuning and Evaluation Setup}
Our procedure followed a carefully controlled protocol:

\begin{itemize}
\item \textbf{Training Configuration}: Models were fine-tuned for 3 epochs using our \ours~dataset.
\item \textbf{Data Format}: Each training sample adheres to the default chat template, with models required to generate both a thinking process (enclosed in \texttt{<think>} and \texttt{</think>} tokens) and a final response. This structured format enables the separation of thinking processes from final responses, facilitating independent evaluation of responses.
\item \textbf{Inference Setup}: Greedy decoding was employed for both benchmarks to ensure reproducibility and fair comparison.
\end{itemize}

\begin{table}[!t]
\small
\centering
\resizebox{\linewidth}{!}{
\begin{tabular}{p{0.98\linewidth}}
\toprule
\textbf{\texttt{Instruction:}} \texttt{Why do a lot of Scientists not believe in God or Satan?} \\
\midrule
\textbf{\texttt{Generated Unstructured \monologue{Monologue}:}} \\
\texttt{Why do many scientists not believe in God or Satan? This is an interesting question. At first glance, it seems like a straightforward matter of faith versus evidence, but there's probably more to it. } \\
\texttt{Okay, so let's break this down. Science is fundamentally about understanding the natural world through observation, experimentation, and evidence. It relies on empirical data and logical reasoning. \textcolor{gray}{...[223 words]...} Wait, perhaps we could consider it from another angle. Many scientists might argue \textcolor{gray}{...[338 words]...} Ah! Now I understand... It's also about the nature of scientific inquiry. Science seeks to explain natural phenomena with natural causes. \textcolor{gray}{...[612 words]...}. } \\
\midrule
\textbf{\texttt{Generated Response:}} \\
\texttt{The reasons why many scientists do not believe in God or Satan can be multifaceted and complex, and it's important to note that individual beliefs can vary widely. Here are some common factors that may contribute to this perspective: } \\
\texttt{1. Scientific Method and Empirical Evidence\textcolor{gray}{...[623 words]...}} \\
\texttt{5. Separation of Science and Religion: Some scientists maintain a distinction between science and religion, viewing them as separate domains that address different types \textcolor{gray}{...[34 words]...}} \\
\texttt{6. Personal Experiences and Backgrounds: Personal backgrounds, upbringing, education, and life experiences play significant roles in shaping one's beliefs. \textcolor{gray}{...[16 words]...}} \\
\texttt{7. Skepticism and Critical Thinking: A strong emphasis on skepticism and critical thinking within the scientific community encourages questioning and challenging \textcolor{gray}{...[22 words]...} } \\
\texttt{It's crucial to remember that not all scientists share this perspective, and many do hold religious beliefs. Additionally, the relationship between science and religion is complex and varies greatly among individuals. \textcolor{gray}{...[568 words]...}} \\
\bottomrule
\end{tabular}
}
\caption{\textbf{Example behavior of Qwen-14B with Unstructured \monologue{Monologue}.}}
\label{tab:example1}
\end{table}
\begin{table}[!t]
\small
\centering
\resizebox{\linewidth}{!}{  
\begin{tabular}{p{0.98\linewidth}}   
\toprule  
\textbf{\texttt{Instruction:}} \texttt{Why do a lot of Scientists not believe in God or Satan?}  \\
\midrule
\textbf{\texttt{Generated \decomposition{Decomposition} Thought:}}\\
\texttt{1. Problem Scoping \& Understanding  - The core problem here is understanding why many scientists may not believe in the existence of God or Satan. \textcolor{gray}{...[81 words]...} } \\  \texttt{2. Component Identification  - Essential building blocks include:    * The scientific method: How it operates and what it values.    * Empirical evidence: Its importance \textcolor{gray}{...[84 words]...}} \\  \texttt{3. Sub-Problem Analysis  - Scientific Method: I need to consider how the scientific method emphasizes testability and falsifiability. This means \textcolor{gray}{...[189 words]...}} \\  \texttt{4. Connection Mapping  - The scientific method and empirical evidence are closely linked; both require claims to be testable \textcolor{gray}{...[652 words]...}} \\  \texttt{5. Integration Strategy  - Combining insights from each component, we see that the preference for empirical evidence and natural \textcolor{gray}{...[812 words]...}} \\  \texttt{6. Solution Validation  - This decomposition covers the main reasons why many scientists might not believe in God or Satan, focusing on the \textcolor{gray}{...[112 words]...}}  \\
\midrule
\textbf{\texttt{Generated Response:}} \\
\texttt{There are several reasons why many scientists may not believe in the existence of God or Satan:} \\  \texttt{1. Scientific Method: Science relies on empirical evidence, observation, and experimentation to understand the natural world. \textcolor{gray}{...[36 words]...}} \\  \texttt{2. Lack of Evidence: Scientists generally require evidence before accepting a claim as true. \textcolor{gray}{...[35 words]...}} \\  \texttt{3. Natural Explanations: Science seeks to explain phenomena through natural causes and processes. \textcolor{gray}{...[42 words]...}} \\  \texttt{4. Philosophical Perspectives: Many scientists adhere to philosophical \textcolor{gray}{...[32 words]...}} \\  \texttt{5. Historical Context: Throughout history, there has been a tension between science and organized religion. \textcolor{gray}{...[48 words]...}} \\  \texttt{It is important to note that not all scientists share the same beliefs, and there are many scientists who do believe in God or Satan. Personal beliefs are subjective and vary among individuals. \textcolor{gray}{...[82 words]...}}   \\
\bottomrule
\caption{\textbf{Example behavior of Qwen-14B with \decomposition{Decomposition} Thought}.}
\label{tab:example2}
\end{tabular}
}
\end{table}

\subsection{Main Results Analysis}

Table~\ref{tab:main} presents comprehensive experimental results comparing different thinking patterns across Qwen2.5 series models of varying sizes. Our analysis reveals several key findings:
\vspace{2em}
\paragraph{Overall Effectiveness} 
All thinking patterns demonstrate substantial improvements over the baseline instruction-response approach across different model sizes. This improvement is particularly evident in larger models, with the performance gap widening as model size increases. For instance, in Qwen2.5-14B, while the baseline achieves a 29.53\% length control win rate in AlpacaEval2, all thinking patterns exceed 40\%, with the best performing pattern reaching 44.68\%. This trend suggests a positive correlation between model capacity and the effectiveness of thinking patterns.

\paragraph{Structured vs. Unstructured Thinking}
Our results demonstrate a notable dichotomy between structured and unstructured thinking approaches. The  monologue pattern, representing an unstructured approach, consistently outperforms other patterns across different model sizes (3B, 14B, and 32B), with particularly strong performance in larger models (achieving 66.20\% win rate on Arena-Hard with 14B and 71.60\% with 32B). Conversely, the highly structured decomposition thinking pattern exhibits an inverse relationship with model size - while effective for smaller models, it shows diminishing returns as model size increases, ultimately yielding the lowest performance among all patterns with 32B models (50.80\% win rate on Arena-Hard).

\vspace{2em}

Qualitative analysis of the results in \ref{tab:example1} and \ref{tab:example2} provide insights into this phenomenon. For Qwen-14B with Decomposition thought pattern, the rigid, structured thinking process appears to constrain the model's flexibility, leading to suboptimal responses when faced with diverse instructions. In contrast, Qwen-14B with unstructured monologue pattern demonstrates greater adaptability, enabling comprehensive multi-perspective analysis and generating more well-rounded responses. This suggests that while structured thinking scaffolds may benefit smaller models, larger models appear to perform better with more flexible, unstructured thinking approaches that allow them to fully leverage their enhanced capabilities.

\paragraph{Size-Dependent Pattern Effectiveness} 
The effectiveness of different thinking patterns exhibits clear dependencies on model size. In smaller models (3B-7B), both structured and unstructured patterns show comparable improvements over baseline. The mid-sized model (14B) appears to be a sweet spot where all thinking patterns provide substantial gains. However, in the largest model (32B), we observe a clear advantage for unstructured thinking, with monologue significantly outperforming structured approaches. This size-dependent effectiveness suggests that optimal thinking patterns should be selected based on model size.

\paragraph{Stability of Two-Stage Thinking like self-critic Thought} 
The self-critic thinking pattern, implementing a two-stage thinking process of generation and evaluation, demonstrates remarkable stability across model sizes. It consistently performs well, never falling to the bottom, and occasionally achieves the best performance (e.g., 36.42\% length control win rate with 7B in AlpacaEval 2). This stability suggests that the evaluate-and-improve iterative thinking paradigm might be a universally beneficial approach, regardless of model size. Furthermore, it maintains competitive performance even in 32B model (71.40\% win rate on Arena-Hard), indicating its effectiveness scales well with model size.

\section{Conclusion}
In this work, we present a comprehensive analysis of how different thinking patterns impact the performance of language models across various model sizes. Through extensive experiments with our \ours~dataset, we discovered that smaller models (<30B parameters) could benefit from structured thinking patterns, while larger models (32B parameters) perform better with unstructured approaches. Notably, the unstructured monologue thinking pattern showed consistent effectiveness across most model sizes.
The release of our Thinking Pattern-21k dataset, along with associated checkpoints and training logs, provides valuable resources for the research community. Our findings open new avenues for research into the relationship between model sizes and thinking patterns, contributing to more efficient approaches in reasoning language model development.

\paragraph{Limitations}

While our work demonstrates promising results in enhancing language models' reasoning capabilities, we acknowledge several limitations. First and foremost, our dataset is exclusively constructed in English, which potentially limits its applicability to multilingual scenarios and may not capture reasoning patterns specific to other languages and cultures. This monolingual focus could result in biased performance when models are applied to non-English tasks or cross-lingual reasoning scenarios.

Future work could address these limitations by extending the dataset to include multiple languages and scaling up the dataset size to cover a broader range of reasoning scenarios.

\paragraph{Acknowledgments}
This work is funded in part by the HKUST Startup Fund (R9911), Theme-based Research Scheme
grant (No.T45-205/21-N) and the InnoHK funding for Hong Kong Generative AI Research and
Development Center, Hong Kong SAR.


\bibliography{custom, jm}
\clearpage
\appendix


\onecolumn
\section{Appendix}
\label{sec:appendix}

\begin{figure*}[!hb]  
  \centering
  \includegraphics[width=0.8\textwidth]{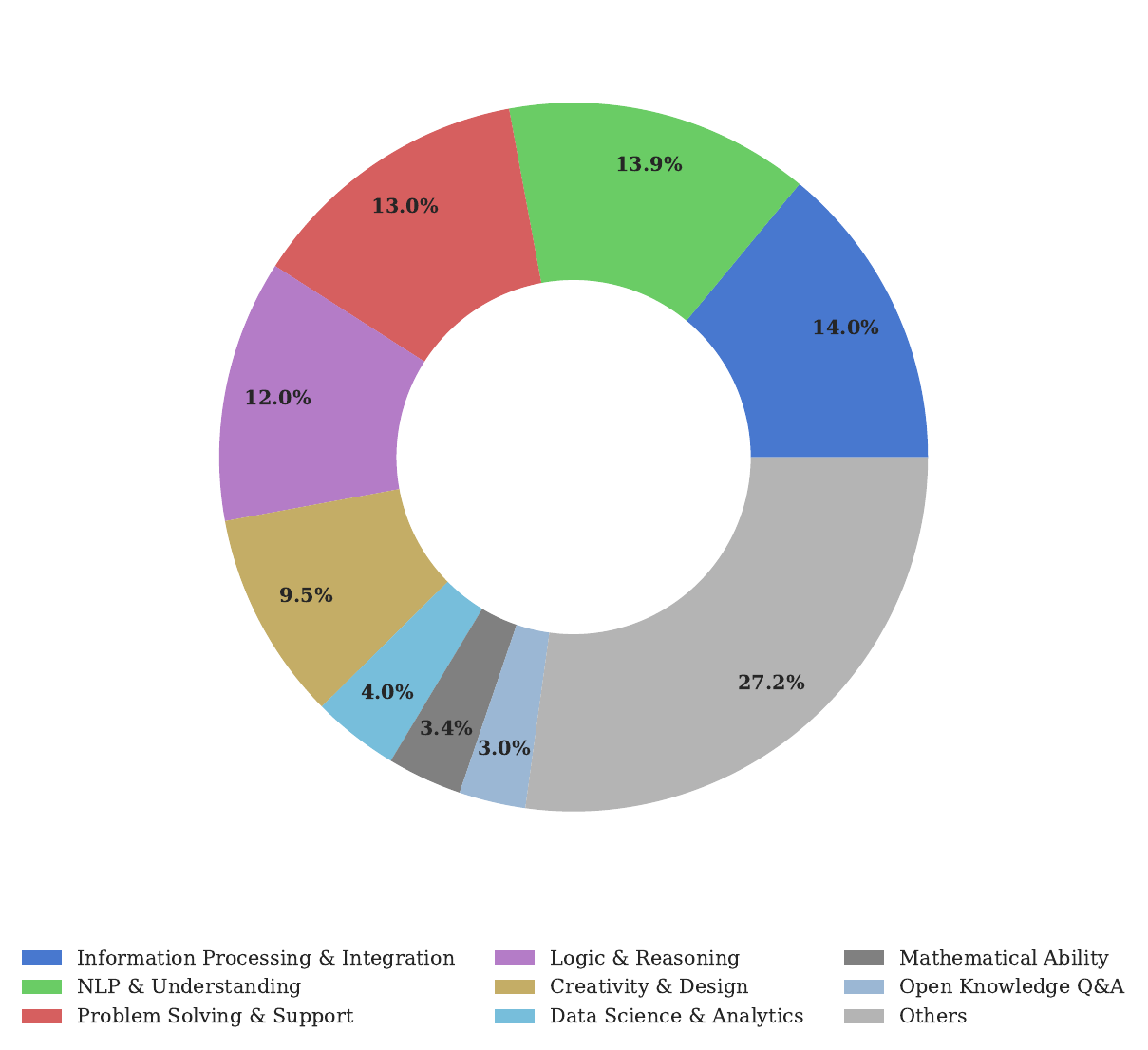}
  \caption{\textbf{Distribution of instructions categories in our dataset.}}
  \label{fig:categories}
\end{figure*}

\begin{table*}[t]
\begin{tcolorbox}
\begin{lstlisting}[
  language=Python,
  breaklines=true,
  breakatwhitespace=true,
  basicstyle=\ttfamily\small
]
SYSTEM_PROMPT: str = """When presented with a question, simulate the internal thought process of an insightful thinker, while naturally steering toward the given answer. Requirements:

1. Thought Characteristics:
- Use first-person monologue form
- Demonstrate authentic thinking processes, including:
  * Initial confusion and questions
  * Progressive analysis
  * Moments of insight
  * Self-correction
  * Multi-perspective thinking
- Use conversational, natural expression
- Show appropriate hesitation and uncertainty

2. Levels of Thinking:
- Problem Understanding: Clarify problem nature and key elements
- Association: Connect with known information and experience
- Multi-dimensional Analysis: Consider multiple possible solution paths
- Comparative Evaluation: Assess pros and cons of different approaches
- Deep Integration: Merge scattered thoughts into coherent understanding
- Practical Application: Consider real-world implementation


3. Expression Style:
"Hmm... let me think about this..."
"This is an interesting question..."
"Wait, perhaps we could consider it from..."
"Ah! Now I understand..."
"No, no, it should be..."
"Connecting these ideas together..."

Output Format:
[Complete internal monologue showing natural progression from confusion to insight]

Please generate an unstructured monologue following these requirements. The monologue should flow naturally and reflect the cognitive development, while  naturally steering toward the given answer."""

USER_PROMPT: str = """
Question: {prompt}
Given Answer: {answer}
Your complete monologue:
"""
\end{lstlisting}
\end{tcolorbox}
\caption{The prompt template for \monologue{monologue} thinking annotation.}
\end{table*}

\begin{table*}[t]
\begin{tcolorbox}
\begin{lstlisting}[
  language=Python,
  breaklines=true,
  breakatwhitespace=true,
  basicstyle=\ttfamily\small
]
SYSTEM_PROMPT: str = """Please act as a systematic decomposition reasoner tasked with breaking down complex problems and generating detailed rationales while naturally steering toward the given answer. Requirements:

Follow this systematic decomposition framework:

1. Problem Scoping & Understanding
- What is the core problem we need to solve?
- What are the key components of this problem?
- What type of solution are we looking for?


2. Component Identification
- What are the essential building blocks of this problem?
- How can we break this down into smaller, manageable parts?
- Which components are independent vs interconnected?


3. Sub-Problem Analysis
- For each component:
  * What specific aspects need to be considered?
  * What approach would work best for this part?
  * What constraints or requirements apply?


4. Connection Mapping
- How do these components interact with each other?
- What are the dependencies between parts?
- What is the logical sequence for solving these parts?


5. Integration Strategy
- How can we combine solutions from different components?
- What potential conflicts might arise during integration?
- How do we ensure consistency across components?

6. Solution Validation
- Does our decomposition cover all aspects of the problem?
- Are there any gaps in our component solutions?
- How well do the parts work together as a whole?



Ouptut Format:
{Your complete decomposition thought}

Please generate a complete internal  decomposition thought following these requirements, while naturally steering toward the given answer."""
USER_PROMPT: str = """
Question: {prompt}
Given Answer: {answer}
Your complete decomposition thought:
"""
\end{lstlisting}
\end{tcolorbox}
\caption{The prompt template for \decomposition{decomposition} thinking annotation.}
\end{table*}

\begin{table*}[t]
\begin{tcolorbox}
\begin{lstlisting}[
  language=Python,
  breaklines=true,
  breakatwhitespace=true,
  basicstyle=\ttfamily\small
]
SYSTEM_PROMPT: str = """Please act as a mentor skilled in the Socratic method of questioning, guiding learners to think deeply and explore problems through progressive questioning while naturally steering toward the given answer. Requirements:

1. Dialogue Format:
- Use "Mentor" and "Learner" to mark the dialogue
- Mentor asks open-ended questions, avoiding yes/no questions
- Learner provides answers that may include initial thoughts, incomplete views, or insights requiring further exploration
- Mentor continues with deeper questions based on learner's responses
- Both parties use "we," "let's," etc., to demonstrate collaborative exploration
- Avoid using the word "you", "I" and other one-directional expressions

2. Questioning Strategies:
- Start with basic concepts and known information
- Guide learners to question assumptions
- Encourage multi-perspective thinking
- Request justification and examples
- Explore cause-and-effect relationships
- Consider different viewpoints
- Identify potential issues
- Summarize key findings

3. Dialogue Progression:
- Beginning: Clarify the problem and basic concepts
- Development: Deep exploration, revealing complexity
- Challenge: Present counterexamples or alternative viewpoints
- Synthesis: Integrate different perspectives for more complete understanding
- Application: Discuss practical applications and implications

4. Levels of Thinking:
- Factual level: What is it?
- Conceptual level: Why?
- Relational level: How does it connect?
- Application level: How can it be used?
- Evaluation level: How can we assess it?
- Innovation level: What new possibilities exist?

5. Finally:
- Summarize key discoveries from the discussion

Ouptut Format:
Mentor: {part of the dialogue}
Learner: {part of the dialogue}

Please generate a complete Socratic dialogue following these requirements. The dialogue should flow naturally and reflect authentic thought processes and cognitive development, while  naturally steering toward the given answer."""
USER_PROMPT: str = """
Question: {prompt}
Given Answer: {answer}
Your complete Socratic dialogue:
"""
\end{lstlisting}
\end{tcolorbox}
\caption{The prompt template for self-ask thinking annotation.}
\end{table*}

\begin{table*}[t]
\begin{tcolorbox}
\begin{lstlisting}[
  language=Python,
  breaklines=true,
  breakatwhitespace=true,
  basicstyle=\ttfamily\small
]
SYSTEM_PROMPT: str = """Please act as a skilled debate facilitator generating an internal dialogue between Proposition and Opposition perspectives on problem-solving, while naturally steering toward the given answer. Requirements:

1. Dialogue Format:
- Use "Proposition" and "Opposition" to mark the dialogue
- Both sides present arguments with evidence and reasoning
- Arguments should be constructive and focused on improving the solution
- Avoid personal attacks or logical fallacies
- Use formal debate language and clear structure

2. Debate Structure:
- Initial Position: Proposition presents an initial solution or approach
- Challenge: Opposition identifies potential weaknesses or limitations
- Defense & Refinement: Proposition addresses concerns and improves solution
- Counter-proposal: Opposition suggests alternative approaches
- Synthesis: Both sides work towards an improved solution

3. Argument Components:
- Clear claims supported by evidence or reasoning
- Specific examples and scenarios
- Consideration of practical constraints
- Analysis of trade-offs and consequences
- Recognition of different contexts and situations

4. Critical Analysis Levels:
- Feasibility: Is it practically achievable?
- Efficiency: Is it the most effective approach?
- Completeness: Are all aspects considered?
- Robustness: Will it work in different scenarios?
- Innovation: Are there better alternatives?

5. Debate Progression:
- Start with fundamental aspects
- Gradually explore deeper complexities
- Consider edge cases and exceptions
- Examine implementation challenges
- Reach refined conclusions

6. Final Synthesis:
- Summarize key points of agreement
- Highlight improvements to original solution
- Note remaining considerations or trade-offs

Output Format:
Proposition: {part of the dialogue}
Opposition: {part of the dialogue}

Please generate a complete debate dialogue following these requirements. The debate should demonstrate rigorous analysis and collaborative problem-solving, while naturally steering toward the given answer."""

USER_PROMPT: str = """
Problem to debate: {prompt}
Given Answer: {answer}
Your complete debate dialogue:
"""
\end{lstlisting}
\end{tcolorbox}
\caption{The prompt template for self-debate thinking annotation.}
\end{table*}

\begin{table*}[t]
\begin{tcolorbox}
\begin{lstlisting}[
  language=Python,
  breaklines=true,
  breakatwhitespace=true,
  basicstyle=\ttfamily\small
]
SYSTEM_PROMPT: str = """When analyzing any query or task, please first generate a Draft Response, then generate Critical Comments, while naturally steering toward the given answer. Requirements:

1. Draft Response:
Generate an initial response

2. Critical Comments:
Analyze your draft response by considering:
- Potential weaknesses or gaps
- Logical flaws or inconsistencies
- Missing perspectives or alternatives
- Areas for improvement
- Suggestions for a better version
- Steering toward the given answer

The critical comments should:
- Be specific and actionable
- Reference particular parts of the draft
- Suggest concrete improvements
- Consider different angles or approaches
- Guide towards a more comprehensive solution

Output Format:
Draft Response:
[Your initial complete response to the instruction]

Critical Comments:
[Your analysis of the draft response, highlighting areas for improvement and suggesting specific enhancements]
"""

USER_PROMPT: str = """
Problem: {prompt}
Given Answer: {answer}
Your Draft Response and Comments:
"""
\end{lstlisting}
\end{tcolorbox}
\caption{The prompt template for self-critic thinking annotation.}
\end{table*}

\onecolumn 
\begin{small}
\begin{longtable}{p{0.98\textwidth}}
\toprule
\textbf{\monologue{Monologue} Example}\\
\midrule

\textbf{Instruction:}\\
\texttt{What are the top wildlife safari destinations in Africa?}
\\
\\
\textbf{Thinking:}\\
\texttt{Hmm... let me think about this... Top wildlife safari destinations in Africa... This is an interesting question. Africa is such a vast continent with so many different ecosystems and wildlife reserves. It's hard to pinpoint just a few, but let's try to break it down.} \\ \\ \texttt{First, there's the Serengeti in Tanzania. I remember reading about the Great Migration, where millions of wildebeest, zebras, and other animals move in a massive loop through the Serengeti and the Maasai Mara in Kenya. That must be one of the most spectacular wildlife events on the planet. The sheer number of animals and the predators that follow them, like lions and crocodiles, make it a top destination for sure.} \\ \\ \texttt{Then there's the Maasai Mara itself. It's essentially an extension of the Serengeti ecosystem, but it's in Kenya. The Maasai Mara is famous for its big cats \u2013 lions, leopards, and cheetahs. I think the BBC's \"Big Cat Diary\" was filmed there. The open savannahs make it easier to spot wildlife, which is a big plus for safari-goers.} \\ \\ \texttt{Wait, perhaps we could consider it from another angle... Southern Africa has some incredible spots too. Kruger National Park in South Africa comes to mind. It's one of the largest game reserves in Africa and has a high density of wildlife, including the Big Five \u2013 lions, leopards, rhinos, elephants, and buffalo. Plus, it's very accessible with good infrastructure, which might be important for some travelers.} \\ \\ \texttt{Ah! Now I understand... Botswana's Okavango Delta is another must-mention. It's a unique inland delta where the Okavango River spreads out into the Kalahari Desert, creating a lush, water-rich area teeming with wildlife. The delta's waterways and islands offer a different kind of safari experience, often involving mokoro (dugout canoe) trips. The diversity of habitats there supports a wide range of animals, including large herds of elephants and hippos.} \\ \\ \texttt{No, no, it should be... Namibia's Etosha National Park also deserves a spot on the list. The park's salt pan is so large it can be seen from space, and it attracts a variety of wildlife, especially during the dry season when animals gather around the waterholes. The stark, otherworldly landscape of Etosha is quite different from the lush environments of other top safari destinations.} \\ \\ \texttt{Connecting these ideas together... So, if I were to summarize, the top wildlife safari destinations in Africa would likely include the Serengeti and Maasai Mara for the Great Migration and big cats, Kruger National Park for its accessibility and Big Five, the Okavango Delta for its unique water-based safaris, and Etosha National Park for its dramatic landscapes and wildlife concentrations.} \\ \\ \texttt{Considering real-world implementation... For someone planning a safari, it might come down to what kind of experience they're looking for \u2013 whether it's witnessing the Great Migration, seeing a high density of wildlife, exploring unique landscapes, or having a more accessible and comfortable journey. Each of these destinations offers something special and unforgettable.}
\\
\\
\textbf{Response:}\\
\texttt{Africa is renowned for its incredible wildlife safaris, offering some of the most picturesque and diverse landscapes along with opportunities to see a wide array of animals in their natural habitats. Here are some of the top wildlife safari destinations on the continent:} \\ \\ \texttt{1. **Masai Mara National Reserve, Kenya**} \\ \texttt{    - Known for the Great Migration, where millions of wildebeest, zebras, and gazelles traverse the Mara River.} \\ \texttt{    - Excellent for spotting the \"Big Five\" (lion, leopard, rhino, elephant, and Cape buffalo).} \\ \\ \texttt{2. **Serengeti National Park, Tanzania**} \\ \texttt{    - Famous for its annual migration and vast open plains.} \\ \texttt{    - Home to a diverse range of wildlife, including large predator populations like lions, cheetahs, and hyenas.} \\ \\ \texttt{3. **Kruger National Park, South Africa**} \\ \texttt{    - One of Africa's largest game reserves.} \\ \texttt{    - Offers a great chance of seeing the Big Five and a wide variety of other wildlife.} \\ \texttt{    - Well-developed tourism infrastructure with various accommodations.} \\ \\ \texttt{4. **Okavango Delta, Botswana**} \\ \texttt{    - A unique inland delta providing lush habitats for a wide array of animals.} \\ \texttt{    - Known for boat safaris, walking safaris, and traditional game drives.} \\ \\ \texttt{5. **Chobe National Park, Botswana**} \\ \texttt{    - Famous for its large elephant population.} \\ \texttt{    - The Chobe Riverfront provides excellent opportunities to see wildlife drawn to the water.} \\ \\ \texttt{6. **Etosha National Park, Namibia**} \\ \texttt{    - Features large salt pans and abundant wildlife around waterholes.} \\ \texttt{    - Unique landscapes and opportunities for both daytime and nighttime game viewing.} \\ \\ \texttt{7. **Bwindi Impenetrable National Park, Uganda**} \\ \texttt{    - Renowned for gorilla trekking experiences.} \\ \texttt{    - Dense, mountainous rainforest habitat offering a chance to see other primates and bird species.} \\ \\ \texttt{8. **Hwange National Park, Zimbabwe**} \\ \texttt{    - Known for its large herds of elephants and predators such as lions and cheetahs.} \\ \texttt{    - Diverse habitats ranging from woodlands to open plains.} \\ \\ \texttt{9. **Ngorongoro Crater, Tanzania**} \\ \texttt{    - A UNESCO World Heritage Site with a vast volcanic caldera home to a dense concentration of wildlife.} \\ \texttt{    - Excellent for spotting the Big Five and viewing a variety of bird species.} \\ \\ \texttt{10. **Rwanda's Volcanoes National Park**} \\ \texttt{    - Famous for mountain gorilla trekking.} \\ \texttt{    - Also home to golden monkeys and other primates.} \\ \\ \texttt{Each of these destinations offers unique experiences and opportunities to see some of Africa's most iconic wildlife. Whether you're interested in traditional game drives, boat safaris, walking safaris, or specialized treks, Africa's safari destinations provide a rich and diverse array of options for wildlife enthusiasts.}
\\
\bottomrule
\caption{Concrete example of~\monologue{monologue} thinking in~\ours.}
\end{longtable}
\end{small}
\twocolumn

\onecolumn 
\begin{small}
\begin{longtable}{p{0.98\textwidth}}
\toprule
\textbf{\decomposition{Decomposition} Example}\\
\midrule

\textbf{Instruction:}\\
\texttt{What challenges did the Bohemian ethnic group face as a minority group in the areas they inhabited?}
\\
\\
\textbf{Thinking:}\\
\texttt{1. Problem Scoping \& Understanding} \\ \texttt{- The core problem here is understanding the specific challenges faced by the Bohemian ethnic group as a minority. This involves analyzing historical and socio-cultural factors that contributed to their marginalization.} \\ \texttt{- Key components of this problem include discrimination in various societal sectors, political marginalization, cultural suppression, religious intolerance, and economic deprivation.} \\ \texttt{- The solution we seek is a comprehensive understanding of these challenges in order to appreciate the broader historical context and its consequences for minority groups.} 
\\ \\ \texttt{2. Component Identification} \\ \texttt{- There are a few essential building blocks to consider:} \\ \texttt{  a) Discrimination: Identifying sectors like employment, housing, and education where discrimination occurred.} \\ \texttt{  b) Political marginalization: Understanding exclusion from power and its impacts.} \\ \texttt{  c) Cultural suppression: Evaluating ways in which cultural aspects were suppressed.} \\ \texttt{  d) Religious intolerance: Determining the religious dynamics influencing their treatment.} \\ \texttt{  e) Economic deprivation: Analyzing economic impacts and contributing factors.} \\ \texttt{- These components are interconnected as each can influence the others, yet they can also be examined individually to build a comprehensive view.} \\ \\ \texttt{3. Sub-Problem Analysis} \\ \texttt{- For discrimination, I should look into historical records and testimonies describing treatment in everyday interactions, laws, and societal norms.} \\ \texttt{- Political marginalization requires exploration of governance structures and representation practices.} \\ \texttt{- Cultural suppression involves investigating policies aimed at diminishing minority identities and linguistic rights.} \\ \texttt{- Addressing religious intolerance means understanding the religious landscape of the time and any dominant religious ideologies in opposition to Catholicism.} \\ \texttt{- Examining economic deprivation involves looking at job market availability, educational barriers, and access to economic resources.} \\ \\ \texttt{4. Connection Mapping} \\ \texttt{- The interaction between political marginalization and economic deprivation is crucial as lack of political power can lead to curtailed access to economic resources.} \\ \texttt{- Discrimination connects with cultural suppression since both are often used to enforce societal hierarchies and assimilation into dominant groups.} \\ \texttt{- Religious intolerance can compound other factors by justifying social ostracization or influencing other discrimination forms.} \\ \texttt{- The logical sequence involves understanding cultural and political contexts first as they frame the broader environment for specific discriminations and economic impacts.} \\ \\ \texttt{5. Integration Strategy} \\ \texttt{- Solutions from each component need to be synthesized into a coherent narrative that considers the interplay of social, economic, and political factors.} \\ \texttt{- Potential conflicts might arise in differing interpretations of historical records which need reconciliation through contextual analysis.} \\ \texttt{- Consistency is ensured by cross-verifying historical accounts and drawing parallels with similar minority challenges in other contexts.} \\ \\ \texttt{6. Solution Validation} \\ \texttt{- By covering political, economic, and cultural aspects, and their interconnections, this approach should comprehensively address the challenges faced by the Bohemians.} \\ \texttt{- Considering historical examples and data helps ensure there are no gaps, while also allowing insights into the broader implications of minority status challenges.} \\ \texttt{- The components should work together to illustrate a clear picture of the lived realities and systemic issues the Bohemians encountered.
}
\\
\\
\textbf{Response:}\\
\texttt{The Bohemian ethnic group, also known as Czechs, faced various challenges as a minority group in the areas they inhabited, including:} \\ \\ \texttt{1. Discrimination: The Bohemians often faced discrimination in employment, housing, education, and public accommodations. They were treated as the second-class citizens by the ruling elites who favored the majority groups.} \\ \\ \texttt{2. Political marginalization: The Bohemians were often excluded from political power and representation, leading to a lack of resources for their communities.} \\ \\ \texttt{3. Cultural suppression: The Bohemians' language and cultural traditions were often suppressed, and they were forced to assimilate into the majority culture.} \\ \\ \texttt{4. Religious intolerance: The Bohemians were primarily Catholic, and they faced discrimination and persecution from other religious groups, especially Protestants.} \\ \\ \texttt{5. Economic deprivation: The Bohemians often faced economic deprivation due to their minority status, which limited their access to education and job opportunities.} \\ \\ \texttt{These challenges negatively impacted the Bohemians' quality of life and the development of their communities in the areas they inhabited.
}
\\
\bottomrule
\caption{Concrete example of~\decomposition{decomposition} thinking in~\ours.}
\end{longtable}
\end{small}
\twocolumn

\onecolumn 
\begin{small}
\begin{longtable}{p{0.98\textwidth}}
\toprule
\textbf{\ask{Self-Ask} Example}\\
\midrule

\textbf{Instruction:}\\
\texttt{What are the top wildlife safari destinations in Africa?}
\\
\\
\textbf{Thinking:}\\
\texttt{Let's begin by considering what we mean by wildlife safari destinations. What do we think are the key elements that make a destination suitable for a wildlife safari?}
\\
\texttt{I think a good wildlife safari destination would have a large variety of animals, especially the big ones like lions, elephants, and rhinos. It should also have good infrastructure for tourists, like lodges and guided tours.} \\ \\ \texttt{ Those are excellent points. We mentioned the variety of animals and infrastructure. Why do we think these elements are important for a wildlife safari experience?} \\ \\ \texttt{ The variety of animals is important because it makes the safari more exciting and educational. The infrastructure is important because it ensures that tourists can have a comfortable and safe experience while exploring the wildlife.} \\ \\ \texttt{ Great. Now, considering these elements, which destinations in Africa come to mind as top wildlife safari locations?} \\ \\ \texttt{ I think of places like the Serengeti in Tanzania, the Maasai Mara in Kenya, and Kruger National Park in South Africa. These are often mentioned as top safari destinations.} \\ \\ \texttt{ Interesting choices. What do we know about the Serengeti, Maasai Mara, and Kruger National Park that supports their reputation as top safari destinations?} \\ \\ \texttt{ The Serengeti is famous for the Great Migration, where millions of wildebeest and other animals move across the plains. The Maasai Mara is known for its large populations of big cats like lions and cheetahs. Kruger National Park is one of the largest game reserves in Africa and has a wide variety of animals and well-developed tourist facilities.} \\ \\ \texttt{ Excellent observations. Let's delve deeper into one of these destinations. How does the Great Migration in the Serengeti contribute to its status as a premier safari destination?} \\ \\ \texttt{ The Great Migration is a unique and spectacular event that attracts many tourists. It involves millions of animals moving in search of water and grazing land, which creates dramatic scenes and opportunities to see predators in action.} \\ \\ \texttt{ Indeed, the Great Migration is a significant draw. How might the presence of predators during the migration enhance the safari experience?} \\ \\ \texttt{ Predators like lions, cheetahs, and crocodiles follow the migration and hunt the animals, which can be thrilling for tourists to witness. It adds an element of excitement and shows the raw, natural behavior of wildlife.} \\ \\ \texttt{ Absolutely. Now, let's consider the Maasai Mara. How does its reputation for big cats contribute to its appeal as a safari destination?} \\ \\ \texttt{ The Maasai Mara's high density of big cats means that tourists have a good chance of seeing lions, cheetahs, and leopards. This makes it a popular destination for those interested in observing these predators up close.} \\ \\ \texttt{ Very true. How does the Maasai Mara's landscape and ecosystem support such a high density of big cats?} \\ \\ \texttt{ The Maasai Mara has vast open plains and abundant prey, which provide ideal hunting grounds for big cats. The ecosystem supports a balanced food chain, ensuring that predators and prey thrive.} \\ \\ \texttt{ Excellent. Now, let's turn our attention to Kruger National Park. What aspects of Kruger make it a top choice for safari-goers?} \\ \\ \texttt{ Kruger National Park is known for its size and diversity of wildlife. It has the Big Five\u2014lions, leopards, rhinos, elephants, and buffaloes\u2014and many other species. It also has well-developed infrastructure, including roads, lodges, and guided tours, making it accessible and comfortable for tourists.} \\ \\ \texttt{ Good points. How does the infrastructure in Kruger enhance the overall safari experience?} \\ \\ \texttt{ The infrastructure allows tourists to explore the park safely and comfortably. Guided tours provide expert knowledge about the animals and their behavior, while lodges offer a place to rest and enjoy the surroundings. This makes the experience more enjoyable and educational.} \\ \\ \texttt{ Indeed. Considering all we've discussed, what might be some potential challenges or downsides to these popular safari destinations?} \\ \\ \texttt{ One challenge could be the high number of tourists, which might affect the natural behavior of the animals and the overall experience. There could also be environmental impacts from the infrastructure and human presence.} \\ \\ \texttt{ Very insightful. How might we address these challenges to ensure sustainable and enjoyable safari experiences?} \\ \\ \texttt{ We could implement stricter regulations on the number of tourists and vehicles allowed in the parks. Promoting eco-friendly practices and supporting conservation efforts can help minimize the environmental impact. Educating tourists about responsible behavior can also make a difference.} \\ \\ \texttt{ Excellent suggestions. To summarize, we've identified the Serengeti, Maasai Mara, and Kruger National Park as top wildlife safari destinations due to their diverse wildlife, unique events like the Great Migration, and well-developed infrastructure. We've also considered the importance of sustainable practices to preserve these experiences for the future. How might we apply this understanding to choose a safari destination that aligns with our interests and values?} \\ \\ \texttt{ We can research each destination to see which one offers the experiences we are most interested in, such as witnessing the Great Migration or observing big cats. We can also look for operators and lodges that prioritize sustainability and conservation to ensure our visit has a positive impact.} \\ \\ \texttt{ Excellent. By considering our interests and values, we can make informed choices that enhance our safari experience while supporting the preservation of these incredible destinations. Thank you for this thoughtful exploration.} \\ \\ \texttt{ Thank you for guiding me through this discussion. I've gained a deeper understanding of what makes a top wildlife safari destination and how to choose one responsibly.}
\\
\\
\textbf{Response:}\\
\texttt{Africa is renowned for its incredible wildlife safaris, offering some of the most picturesque and diverse landscapes along with opportunities to see a wide array of animals in their natural habitats. Here are some of the top wildlife safari destinations on the continent:} \\ \\ \texttt{1. **Masai Mara National Reserve, Kenya**} \\ \texttt{    - Known for the Great Migration, where millions of wildebeest, zebras, and gazelles traverse the Mara River.} \\ \texttt{    - Excellent for spotting the \"Big Five\" (lion, leopard, rhino, elephant, and Cape buffalo).} \\ \\ \texttt{2. **Serengeti National Park, Tanzania**} \\ \texttt{    - Famous for its annual migration and vast open plains.} \\ \texttt{    - Home to a diverse range of wildlife, including large predator populations like lions, cheetahs, and hyenas.} \\ \\ \texttt{3. **Kruger National Park, South Africa**} \\ \texttt{    - One of Africa's largest game reserves.} \\ \texttt{    - Offers a great chance of seeing the Big Five and a wide variety of other wildlife.} \\ \texttt{    - Well-developed tourism infrastructure with various accommodations.} \\ \\ \texttt{4. **Okavango Delta, Botswana**} \\ \texttt{    - A unique inland delta providing lush habitats for a wide array of animals.} \\ \texttt{    - Known for boat safaris, walking safaris, and traditional game drives.} \\ \\ \texttt{5. **Chobe National Park, Botswana**} \\ \texttt{    - Famous for its large elephant population.} \\ \texttt{    - The Chobe Riverfront provides excellent opportunities to see wildlife drawn to the water.} \\ \\ \texttt{6. **Etosha National Park, Namibia**} \\ \texttt{    - Features large salt pans and abundant wildlife around waterholes.} \\ \texttt{    - Unique landscapes and opportunities for both daytime and nighttime game viewing.} \\ \\ \texttt{7. **Bwindi Impenetrable National Park, Uganda**} \\ \texttt{    - Renowned for gorilla trekking experiences.} \\ \texttt{    - Dense, mountainous rainforest habitat offering a chance to see other primates and bird species.} \\ \\ \texttt{8. **Hwange National Park, Zimbabwe**} \\ \texttt{    - Known for its large herds of elephants and predators such as lions and cheetahs.} \\ \texttt{    - Diverse habitats ranging from woodlands to open plains.} \\ \\ \texttt{9. **Ngorongoro Crater, Tanzania**} \\ \texttt{    - A UNESCO World Heritage Site with a vast volcanic caldera home to a dense concentration of wildlife.} \\ \texttt{    - Excellent for spotting the Big Five and viewing a variety of bird species.} \\ \\ \texttt{10. **Rwanda's Volcanoes National Park**} \\ \texttt{    - Famous for mountain gorilla trekking.} \\ \texttt{    - Also home to golden monkeys and other primates.} \\ \\ \texttt{Each of these destinations offers unique experiences and opportunities to see some of Africa's most iconic wildlife. Whether you're interested in traditional game drives, boat safaris, walking safaris, or specialized treks, Africa's safari destinations provide a rich and diverse array of options for wildlife enthusiasts.}
\\
\bottomrule
\caption{Concrete example of~self-ask thinking in~\ours.}
\end{longtable}
\end{small}
\twocolumn

\onecolumn 
\begin{small}
\begin{longtable}{p{0.98\textwidth}}
\toprule
\textbf{\debate{Self-Debate} Example}\\
\midrule

\textbf{Instruction:}\\
\texttt{What are the top wildlife safari destinations in Africa?}
\\
\\
\textbf{Thinking:}\\
\texttt{Proposition: One of the top wildlife safari destinations in Africa is the Serengeti National Park in Tanzania. The Serengeti is renowned for its annual migration of over 1.5 million wildebeest and 250,000 zebras, making it a unique and unparalleled wildlife spectacle. The park also offers diverse habitats, including savannahs, woodlands, and riverine forests, which support a wide variety of wildlife species such as lions, elephants, and cheetahs.} \\ \\ \texttt{Opposition: While the Serengeti is indeed a remarkable destination, it is important to consider the potential overcrowding and environmental impact due to its popularity. High tourist numbers can lead to habitat degradation and disturb the natural behavior of wildlife. Additionally, the cost of safaris in the Serengeti can be prohibitive for many travelers, limiting accessibility.} \\ \\ \texttt{Proposition: To address the concerns of overcrowding and environmental impact, the Serengeti has implemented strict regulations on the number of vehicles allowed in certain areas and has designated specific zones for tourism. These measures help to minimize human impact on wildlife and their habitats. Furthermore, there are various budget-friendly options available, such as group tours and camping safaris, which make the Serengeti more accessible to a wider range of travelers.} \\ \\ \texttt{Opposition: Another top wildlife safari destination that should be considered is the Okavango Delta in Botswana. The Okavango Delta is a unique inland delta that provides a diverse range of ecosystems, from permanent marshlands to seasonal floodplains. This diversity supports an incredible array of wildlife, including large populations of elephants, hippos, and crocodiles, as well as numerous bird species. The delta's remote location also means that it experiences less tourist traffic, allowing for a more intimate and exclusive safari experience.} \\ \\ \texttt{Proposition: The Okavango Delta is indeed a fantastic destination, but its remoteness can also pose challenges. Accessing the delta often requires charter flights, which can be expensive and logistically complex. Additionally, the seasonal flooding can limit accessibility to certain areas and affect the availability of wildlife sightings.} \\ \\ \texttt{Opposition: To mitigate these challenges, many safari operators in the Okavango Delta offer all-inclusive packages that cover transportation, accommodation, and guided tours, simplifying the logistics for travelers. The seasonal flooding, while a potential limitation, also creates unique opportunities for water-based safaris, such as mokoro (dugout canoe) excursions, which provide a different perspective on wildlife viewing.} \\ \\ \texttt{Proposition: Another noteworthy destination is Kruger National Park in South Africa. Kruger is one of Africa's largest game reserves and offers an extensive network of roads and accommodations, catering to both self-drive and guided safari experiences. The park is home to the Big Five (lion, leopard, rhinoceros, elephant, and Cape buffalo) and boasts a high density of wildlife, making it an excellent choice for first-time safari-goers.} \\ \\ \texttt{Opposition: While Kruger National Park is highly accessible and offers a range of amenities, its popularity can lead to crowded conditions, especially in peak seasons. This can detract from the wilderness experience and increase the risk of human-wildlife conflicts. Additionally, the park's extensive infrastructure can sometimes give it a less pristine feel compared to more remote destinations.} \\ \\ \texttt{Proposition: To enhance the wilderness experience in Kruger, visitors can opt for private game reserves adjacent to the park, such as Sabi Sands or Timbavati. These reserves offer exclusive access to wildlife-rich areas and often provide more personalized and intimate safari experiences. Furthermore, Kruger's well-developed infrastructure ensures that conservation efforts are well-funded and effective, contributing to the long-term preservation of its diverse ecosystems.} \\ \\ \texttt{Opposition: Considering all the points discussed, it is clear that each destination has its unique strengths and challenges. To synthesize an improved solution, we should recommend a combination of destinations based on travelers' preferences and priorities. For those seeking iconic wildlife spectacles and diverse habitats, the Serengeti is an excellent choice. For a more exclusive and intimate experience, the Okavango Delta is ideal. For accessibility and comprehensive safari options, Kruger National Park stands out.} \\ \\ \texttt{Proposition: Agreed. By recognizing the distinct advantages of each destination and considering travelers' individual needs, we can provide a more tailored and enriching safari experience. It is also important to emphasize sustainable tourism practices across all destinations to ensure the long-term preservation of Africa's incredible wildlife and natural landscapes.} \\ \\ \texttt{Opposition: In conclusion, the top wildlife safari destinations in Africa include the Serengeti National Park, the Okavango Delta, and Kruger National Park. Each offers unique experiences and opportunities for wildlife viewing. By promoting sustainable tourism and considering travelers' preferences, we can enhance the overall safari experience while ensuring the protection of these precious ecosystems.} \\ \\ \texttt{Proposition: Well said. By combining our insights and focusing on sustainable practices, we can help travelers make informed choices and contribute to the conservation of Africa's wildlife heritage.
}
\\
\\
\textbf{Response:}\\
\texttt{Africa is renowned for its incredible wildlife safaris, offering some of the most picturesque and diverse landscapes along with opportunities to see a wide array of animals in their natural habitats. Here are some of the top wildlife safari destinations on the continent:} \\ \\ \texttt{1. **Masai Mara National Reserve, Kenya**} \\ \texttt{    - Known for the Great Migration, where millions of wildebeest, zebras, and gazelles traverse the Mara River.} \\ \texttt{    - Excellent for spotting the \"Big Five\" (lion, leopard, rhino, elephant, and Cape buffalo).} \\ \\ \texttt{2. **Serengeti National Park, Tanzania**} \\ \texttt{    - Famous for its annual migration and vast open plains.} \\ \texttt{    - Home to a diverse range of wildlife, including large predator populations like lions, cheetahs, and hyenas.} \\ \\ \texttt{3. **Kruger National Park, South Africa**} \\ \texttt{    - One of Africa's largest game reserves.} \\ \texttt{    - Offers a great chance of seeing the Big Five and a wide variety of other wildlife.} \\ \texttt{    - Well-developed tourism infrastructure with various accommodations.} \\ \\ \texttt{4. **Okavango Delta, Botswana**} \\ \texttt{    - A unique inland delta providing lush habitats for a wide array of animals.} \\ \texttt{    - Known for boat safaris, walking safaris, and traditional game drives.} \\ \\ \texttt{5. **Chobe National Park, Botswana**} \\ \texttt{    - Famous for its large elephant population.} \\ \texttt{    - The Chobe Riverfront provides excellent opportunities to see wildlife drawn to the water.} \\ \\ \texttt{6. **Etosha National Park, Namibia**} \\ \texttt{    - Features large salt pans and abundant wildlife around waterholes.} \\ \texttt{    - Unique landscapes and opportunities for both daytime and nighttime game viewing.} \\ \\ \texttt{7. **Bwindi Impenetrable National Park, Uganda**} \\ \texttt{    - Renowned for gorilla trekking experiences.} \\ \texttt{    - Dense, mountainous rainforest habitat offering a chance to see other primates and bird species.} \\ \\ \texttt{8. **Hwange National Park, Zimbabwe**} \\ \texttt{    - Known for its large herds of elephants and predators such as lions and cheetahs.} \\ \texttt{    - Diverse habitats ranging from woodlands to open plains.} \\ \\ \texttt{9. **Ngorongoro Crater, Tanzania**} \\ \texttt{    - A UNESCO World Heritage Site with a vast volcanic caldera home to a dense concentration of wildlife.} \\ \texttt{    - Excellent for spotting the Big Five and viewing a variety of bird species.} \\ \\ \texttt{10. **Rwanda's Volcanoes National Park**} \\ \texttt{    - Famous for mountain gorilla trekking.} \\ \texttt{    - Also home to golden monkeys and other primates.} \\ \\ \texttt{Each of these destinations offers unique experiences and opportunities to see some of Africa's most iconic wildlife. Whether you're interested in traditional game drives, boat safaris, walking safaris, or specialized treks, Africa's safari destinations provide a rich and diverse array of options for wildlife enthusiasts.}
\texttt{}
\\
\bottomrule
\caption{Concrete example of~self-debate thinking in~\ours.}
\end{longtable}
\end{small}
\twocolumn

\onecolumn 
\begin{small}
\begin{longtable}{p{0.98\textwidth}}
\toprule
\textbf{\critic{Self-Critic} Example}\\
\midrule

\textbf{Instruction:}\\
\texttt{What are the top wildlife safari destinations in Africa?}
\\
\\
\textbf{Thinking:}\\
\texttt{Draft Response:} \\ \texttt{ Africa is renowned for its incredible wildlife and offers numerous safari destinations where visitors can experience the continent's natural beauty and diverse animal species. Here are some of the top wildlife safari destinations in Africa:} \\ \\ \texttt{1. Serengeti National Park, Tanzania: Famous for the Great Migration, where millions of wildebeest, zebras, and gazelles migrate in search of fresh grazing. The park is also home to the Big Five (lion, leopard, elephant, buffalo, and rhino).} \\ \\ \texttt{2. Maasai Mara National Reserve, Kenya: Known for its abundant wildlife and the annual migration of wildebeest and zebras. The reserve offers excellent opportunities to see the Big Five and other species like cheetahs and hyenas.} \\ \\ \texttt{3. Kruger National Park, South Africa: One of Africa's largest game reserves, Kruger is home to a vast array of wildlife, including the Big Five. The park offers a range of accommodation options and well-maintained infrastructure for self-drive safaris.} \\ \\ \texttt{4. Chobe National Park, Botswana: Famous for its large elephant population, Chobe also offers boat safaris on the Chobe River, where visitors can see hippos, crocodiles, and a variety of bird species.} \\ \\ \texttt{5. Etosha National Park, Namibia: Known for its vast salt pan, which attracts a diverse range of wildlife, including elephants, lions, and rhinos. The park's waterholes provide excellent game-viewing opportunities.} \\ \\ \texttt{6. Hwange National Park, Zimbabwe: Home to one of the largest elephant populations in Africa, Hwange also boasts a rich diversity of other wildlife, including lions, leopards, and wild dogs.} \\ \\ \texttt{7. Okavango Delta, Botswana: A unique inland delta that supports a rich variety of wildlife. The delta's waterways are ideal for mokoro (dugout canoe) safaris, offering a different perspective on the animals and their habitats.} \\ \\ \texttt{8. Ngorongoro Crater, Tanzania: A UNESCO World Heritage Site, the crater is a natural enclosure for a wide variety of wildlife, including the Big Five. The crater's floor is teeming with animals, making it a prime safari destination.} \\ \\ \texttt{Critical Comments:} \\ \texttt{ - The draft response provides a good overview of some of the top wildlife safari destinations in Africa, but it could benefit from more detailed descriptions of each location.} \\ \texttt{- The response could include information about the best times to visit each destination to enhance the reader's planning.} \\ \texttt{- Mentioning specific activities or unique experiences available at each destination would add value.} \\ \texttt{- Including a brief note on conservation efforts or the importance of sustainable tourism in these areas could provide a more comprehensive perspective.} \\ \texttt{- The response could be improved by adding a conclusion that summarizes the key points and emphasizes the diversity of safari experiences available in Africa.} \\ \\ \texttt{Suggestions for Improvement:} \\ \texttt{1. Add more detailed descriptions of each destination, highlighting unique features and specific wildlife species that can be seen.} \\ \texttt{2. Include information about the best times to visit each destination to help readers plan their trips.} \\ \texttt{3. Mention specific activities or unique experiences available at each location, such as hot air balloon rides, walking safaris, or night drives.} \\ \texttt{4. Incorporate a brief discussion on conservation efforts and the importance of sustainable tourism in these areas.} \\ \texttt{5. Conclude with a summary that emphasizes the diversity and richness of safari experiences in Africa.} 
\\
\\
\textbf{Response:}\\
\texttt{Africa is renowned for its incredible wildlife safaris, offering some of the most picturesque and diverse landscapes along with opportunities to see a wide array of animals in their natural habitats. Here are some of the top wildlife safari destinations on the continent:} \\ \\ \texttt{1. **Masai Mara National Reserve, Kenya**} \\ \texttt{    - Known for the Great Migration, where millions of wildebeest, zebras, and gazelles traverse the Mara River.} \\ \texttt{    - Excellent for spotting the \"Big Five\" (lion, leopard, rhino, elephant, and Cape buffalo).} \\ \\ \texttt{2. **Serengeti National Park, Tanzania**} \\ \texttt{    - Famous for its annual migration and vast open plains.} \\ \texttt{    - Home to a diverse range of wildlife, including large predator populations like lions, cheetahs, and hyenas.} \\ \\ \texttt{3. **Kruger National Park, South Africa**} \\ \texttt{    - One of Africa's largest game reserves.} \\ \texttt{    - Offers a great chance of seeing the Big Five and a wide variety of other wildlife.} \\ \texttt{    - Well-developed tourism infrastructure with various accommodations.} \\ \\ \texttt{4. **Okavango Delta, Botswana**} \\ \texttt{    - A unique inland delta providing lush habitats for a wide array of animals.} \\ \texttt{    - Known for boat safaris, walking safaris, and traditional game drives.} \\ \\ \texttt{5. **Chobe National Park, Botswana**} \\ \texttt{    - Famous for its large elephant population.} \\ \texttt{    - The Chobe Riverfront provides excellent opportunities to see wildlife drawn to the water.} \\ \\ \texttt{6. **Etosha National Park, Namibia**} \\ \texttt{    - Features large salt pans and abundant wildlife around waterholes.} \\ \texttt{    - Unique landscapes and opportunities for both daytime and nighttime game viewing.} \\ \\ \texttt{7. **Bwindi Impenetrable National Park, Uganda**} \\ \texttt{    - Renowned for gorilla trekking experiences.} \\ \texttt{    - Dense, mountainous rainforest habitat offering a chance to see other primates and bird species.} \\ \\ \texttt{8. **Hwange National Park, Zimbabwe**} \\ \texttt{    - Known for its large herds of elephants and predators such as lions and cheetahs.} \\ \texttt{    - Diverse habitats ranging from woodlands to open plains.} \\ \\ \texttt{9. **Ngorongoro Crater, Tanzania**} \\ \texttt{    - A UNESCO World Heritage Site with a vast volcanic caldera home to a dense concentration of wildlife.} \\ \texttt{    - Excellent for spotting the Big Five and viewing a variety of bird species.} \\ \\ \texttt{10. **Rwanda's Volcanoes National Park**} \\ \texttt{    - Famous for mountain gorilla trekking.} \\ \texttt{    - Also home to golden monkeys and other primates.} \\ \\ \texttt{Each of these destinations offers unique experiences and opportunities to see some of Africa's most iconic wildlife. Whether you're interested in traditional game drives, boat safaris, walking safaris, or specialized treks, Africa's safari destinations provide a rich and diverse array of options for wildlife enthusiasts.}
\texttt{}
\\
\bottomrule
\caption{Concrete example of~self-critic thinking in~\ours.}
\end{longtable}
\end{small}
\twocolumn

\end{document}